# Targeted-BEHRT: Deep learning for observational causal inference on longitudinal electronic health records

Shishir Rao, Mohammad Mamouei, Gholamreza Salimi-Khorshidi, Yikuan Li, Rema Ramakrishnan, Abdelaali Hassaine, Dexter Canoy, Kazem Rahimi

*Abstract*— **Observational causal inference is useful for decision making in medicine when randomized clinical trials (RCT) are infeasible or non-generalizable. However, traditional approaches fail to deliver unconfounded causal conclusions in practice. The rise of "doubly robust" non-parametric tools coupled with the growth of deep learning for capturing rich representations of multimodal data, offers a unique opportunity to develop and test such models for causal inference on comprehensive electronic health records (EHR). In this paper, we investigate causal modelling of an RCT-established null causal association: the effect of antihypertensive use on incident cancer risk. We develop a dataset for our observational study and a Transformer-based model, Targeted BEHRT coupled with doubly robust estimation, we estimate average risk ratio (RR). We compare our model to benchmark statistical and deep learning models for causal inference in multiple experiments on semi-synthetic derivations of our dataset with various types and intensities of confounding. In order to further test the reliability of our approach, we test our model on situations of limited data. We find that our model provides more accurate estimates of RR (least sum absolute error from ground truth) compared to benchmarks for risk ratio estimation on high-dimensional EHR across experiments. Finally, we apply our model to investigate the original case study: antihypertensives' effect on cancer and demonstrate that our model generally captures the validated null association.**

*Index Terms*—**deep learning, electronic health records, causal inference**

## I. Introduction

THE growing availability of routinely collected, administrative clinical records databases with linked Electronic Health Records (EHR) with numerous variables describing large numbers of individuals on a population level provide an encouraging opportunity to conduct observational studies in the medical domain[1]. These administrative EHR datasets such as Clinical Practice Research Datalink (CPRD) in the UK offer a multitude of temporal and static variables in addition to data linkages to other datasets[2] ultimately providing substantive variables for confounding adjustment.

Also, recently deep learning has shown promise in high-dimensional data modelling for prediction tasks in domains of computer vision, natural language processing, and models like BEHRT have shown superior predictive performance in EHR based tasks such as incident disease prediction[3]–[5]. With the recent success of deep learning in predictive modelling and advantages of automatic feature extraction[5], we argue that deep learning for causal inference coupled with utilization of comprehensive EHR shall allow for better confounding adjustment and more accurate estimation of RR than traditional methods.

As a case study, we look at the Randomized Clinical Trial (RCT) investigations of the effect of antihypertensives on cancer. Deemed as null by numerous trials[6], the effect is investigated in the following way: researchers randomly treat patients with different classes of antihypertensives – e.g. some treated with ACE inhibitors and others, beta-blockers – and then compare incidence of cancer between exposure groups. This comparison is often measured as empirical risk ratio (RR), the proportion of cancer in one exposure group divided by the same in the other. Since patients are randomized, trials offer unconfounded estimates of effect on cancer.

However, where an RCT does not exist, is unfeasible or fails to generalize, decision makers will have to draw upon evidence from observational data.[7]–[9] In this observational capacity, adjustment of confounding is a necessity, and insufficient adjustment can result in biased conclusions. Conventionally, by fully adjusting confounding variables in a predictive model, we say the causal effect is identifiable ("no hidden confounding" setting).

In this work, we develop a model, Targeted-BEHRT (T-BEHRT) to conduct observational causal inference in EHR in routine, clinical data available in CPRD and investigate the aforementioned case study: the effect of antihypertensives on cancer. We form our observational dataset by including patients in the UK taking different classes of antihypertensives and investigate risk of cancer. Since counterfactual outcomes are

This research was supported by grants from the Oxford Martin School (OMS), University of Oxford, National Institute for Health Research (NIHR) Oxford Biomedical Research Centre, British Heart Foundation, and UKRI's Global Challenge Research Fund. The views expressed are those of the authors and not necessarily those of the OMS, the BHF, the GRCF, the NIHR or the Department of Health and Social Care.

Rao S., Mamouei, M., Li Y., Ramakrishnan R., Hassaine A., Canoy D., Salimi-Khorshidi G., and Rahimi K. are with the Nuffield Department of Women's & Reproductive Health, University of Oxford.

Rao S. is the corresponding author. (e-mail: shishir.rao@wrh.ox.ac.uk)

.

missing in our observational dataset, ground truth RR is inaccessible in our study of this association thereby making model comparisons difficult. Thus, we first construct semi-synthetic derivations of our observational dataset, and then apply our model against statistical and deep learning benchmarks on several experiments to identify the model with best RR estimation. Second, to test our model in situations of limited data, we demonstrate the utility of T-BEHRT compared to other models in finite-sample estimation experiments.

Lastly, after validating our model on semi-synthetic derivations of routine clinical observational data, we apply our model to investigate the effect of ACEIs on cancer relative to other drug classes. Where traditional statistical models have demonstrated conflicting results, these associations have been deemed null in numerous RCTs[10], [11] with narrow confidence intervals, across a wide range of patient groups, for multiple cancer subtypes.

### *A. Background*

Traditionally, statistical models like logistic regression have been staple models for observational causal inference in epidemiology. However, these models have known limitations[12]. First, they require careful manual feature engineering – useful for modelling known confounders but impractical for those unknown or interacting variables. Furthermore, due to the nature of estimating causality with finite samples of populations, these models are susceptible to finite-sample bias in causal estimation[13]. One solution proposed is to adjust solely for the variables that are associated to exposure by propensity score modelling[14]; however, propensity score-based methods require correct specification of the exposure prediction model, which may not be guaranteed[15]. More recent work in semi-parametric estimation theory – namely "doubly robust" estimation theory – circumvents problems of these modelling strategies. These doubly robust estimators rely on the consistency of either prediction of propensity score or prediction of outcome to produce unbiased causal effect estimates[16], [17], and examples such as Targeted Maximum Likelihood Estimation (TMLE) and derivatives such as the Cross Validated TMLE (CV-TMLE) have been prolifically used to explore causal inference problems of average treatment effect (ATE).

More recently, there have been advances in deep learning for causal inference. Models like TARNET, Dragonnet, CEVAE, and others have been tested on synthetic and semi-synthetic derivations of static tabular data.[7], [18]–[20] However, many of these datasets have limited number of variables and complexity.[21], [22] Furthermore, these models have not been specifically tested in data and experimental settings involving routine multivariate, comprehensive EHR data. And even though deep learning can incorporate multimodal variables, few approaches firstly model both temporal and static variables for causal inference and secondly, develop environments to objectively test proposed solutions against benchmarks. Lastly, the considerable literature of deep learning for causal inference investigates ATE, conditional ATE, and Individualised Treatment Effect (ITE) almost exclusively; methods have rarely been evaluated for robustness in RR – a metric preferred by clinicians since RR captures risk relative to baseline risk (i.e. risk in the control cohort).

Our model, T-BEHRT combines advances in deep EHR modelling and causal inference by utilising an expanded BEHRT for modelling temporal and static EHR data. Furthermore, T-BEHRT models propensity thus allowing for estimation correction with CV-TMLE in order to mitigate finite-sample estimation bias. Lastly, we model auxiliary unsupervised tasks in tandem with learning the causal objective to ultimately aid causal adjustment processes. Previous works[23]–[25] demonstrate auxiliary unsupervised learning 1) adds an additional inductive bias ultimately improving generalizability and 2) helps to learn representations shared or beneficial for the main task – confounding adjustment and causal inference. We hypothesize that incorporation of unsupervised representation learning alongside propensity modelling will help provide more accurate estimates of RR.

## II. Methods

### *A. Dataset and patient selection*

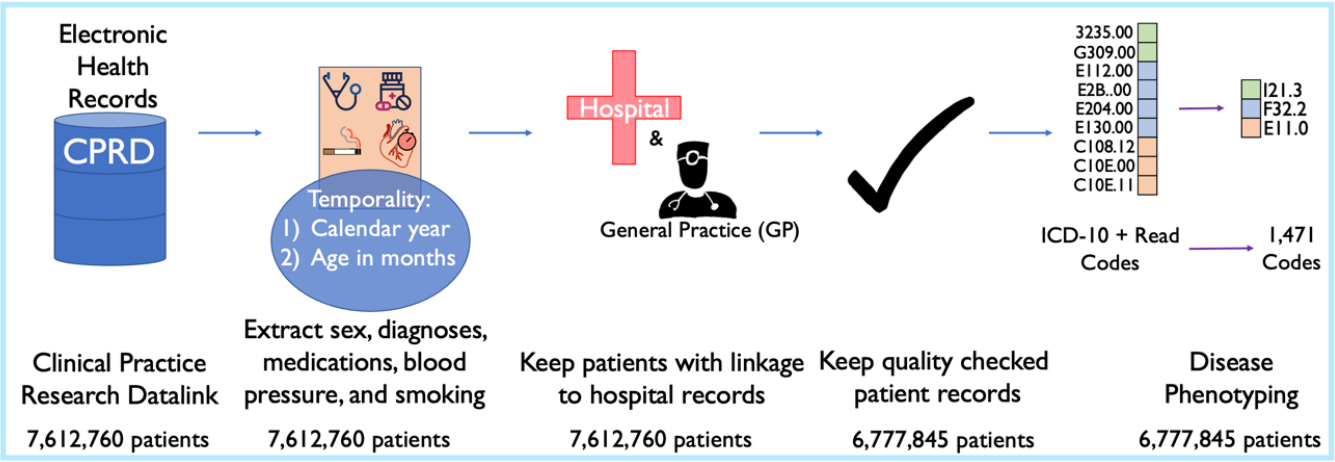

Fig. 1. Representation learning data selection pipeline. We use Clinical Practice Research Datalink (CPRD) and extract diagnoses, medications, blood pressure, smoking, and sex records. We homogenize codes from ICD-10 and Read to one format. Unmapped Read codes were kept for completeness.

For our investigations, we used a cut from CPRD, which has been described previously[5]: It entails records from 1 January 1985 up to 31 Dec 2015 and is linked to national administrative databases including hospitalisations (Hospital Episode Statistics, or HES), and death registration (from Office of National Statistics).

The data for the investigations was restricted to patients in the database who met the following criteria: (1) registered with the general practice for at least 12 months, (2) aged ≥16 years at registration, (3) registered with the practice considered providing 'up-to-standard' data to CPRD, (4) individual data marked by CPRD to be of 'acceptable' quality for research purposes (as determined by CPRD), and (5) registered with a practice that provided consent for linking the data with national databases for hospitalisations and death registry. We mapped diagnoses and medication codes to a homogenized format for machine readability. This led to a dataset of 6,777,845 patients, which was used for general representation learning (shown in **Fig. 1**) for deep learning models.

For our causal inference investigation (i.e., investigating the effect of antihypertensive on incident cancer), a dataset containing five subpopulations had to be selected – one for each class of antihypertensives: ACEIs, diuretics, CCBs, Beta Blockers (BBs), and Angiotensin II Receptor Blockers (ARBs).

Patients were selected in one of these groups based on first class of antihypertensive medications recorded before 2009 and if free of cancer report before this first prescription; the year 2009 was chosen conveniently to have sufficient 'follow-up' time for the occurrence of potential cancers. The date of this first prescription was defined as 'baseline' (a date between 1985 and 31 December 2008). Patients were then followed up from baseline until cancer diagnosis (including cancer diagnoses as cause of death) or end of five-year follow-up period. The learning period included the entire patients' medical records up to a random point between 6 and 12 months before baseline; this is to account for any potential inaccuracies in timing of prescription (or decision to prescribe) and to avoid possibility of antihypertensive prescription itself influencing the model training. "CPRD Product codes" are used for identifying classes of antihypertensives and the set of codes were obtained from a dataset published by University of Bristol[26]. Codes for cancer are found in **Supplementary Table 1** and derived from clinically established publication of codes[27].

### *B. Semi-synthetic data derivation*

Data generation of sequential, temporal variables is a difficult task, and currently, there is no medically validated method of generating realistic EHR medical history. Thus, we utilised given medical history in observational data to exclusively simulate binary factual and counterfactual outcomes.

Inspired by other semi-synthetic data simulations[20], [28], intuitively, we first modelled the association between a medical history variable $Z_i$ (e.g. some diagnosis/medication) and exposure $T_i$ with the empirical propensity in the dataset: $\lambda_i = P(T_i = 1|Z_i)$. If associated to an exposure ($\lambda_i \neq 0.5$), we generated the conditional outcomes, ${Y^{T=1}}_i$ and ${Y^{T=0}}_i$ as a function of $\lambda_i$ and exposure $T_i = 1$ and $T_i = 0$ respectively. In this way, semi-synthetic outcomes arose from an association between $Z_i$ and exposure and $Z_i$ and the outcome. Thus, the relationship between exposure and outcome is confounded by $Z_i$. While the empirical RR – the proportion of the outcome in one exposure group divided by the same in the other – would yield confounded causal conclusions, effectively adjusting for $Z_i$ would yield identifiable causal association between exposure and outcome.

In addition, to test model adjustment potential in situations of varying confounding intensity, we weighted the contribution of the confounding with a $\beta$ factor: greater the $\beta$ implies greater the confounding. More details of the semi-synthetic data generative process and functions modelled are in **Supplementary Methods: Semi-synthetic data generation.**

In our work, we present investigations in semi-synthetic data utilising two forms of confounders: *persisting* and *transient* confounding. We define *persisting* confounding as confounders that are assigned at birth and persist through one's life course; e.g. ethnicity, sex, genes, and other variables assigned at birth that associate to variables later in age. We define *transient* confounding as confounders that manifest at a point or period of one's life effecting events downstream in time; e.g. disease diagnoses, age itself, prescriptions, and other variables not assigned at birth. These two distinctions of confounding are presented in this work because they naturally capture prevalent forms of confounding seen in population health databases.[29] A visual depiction can be found in **Fig. 2.**

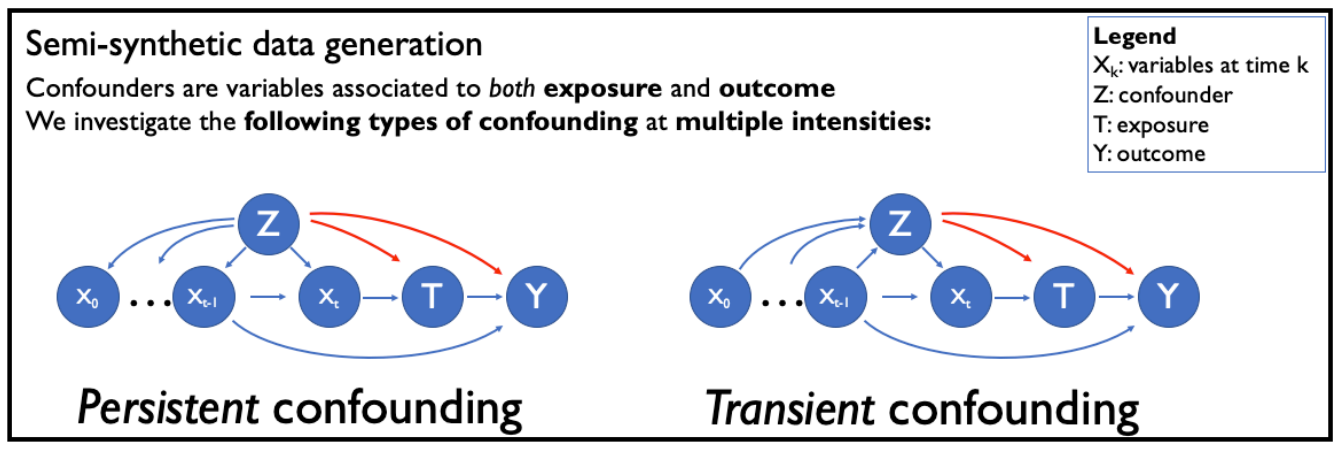

Fig. 2. Confounding forms investigated in semi-synthetic data investigations. We pursue semi-synthetic data generation with confounding variable Z. The two types we define are persistent and transient confounding. The first is a potential confounder assigned at birth and effect's all variables in one's life course. Transient confounding is a potential confounder manifesting at a certain point of time after birth and only confounding downstream associations.

From our observational dataset, we investigated two exposure groups – ACEIs and Diuretics and noticed female sex was associated to the Diuretics exposure status and thus, chose it to be a persistent confounder and generated conditional outcomes. For another pair of exposures: ARBs and CCBs, we identified association of incidence of at least one of heart failure, hypertension, ischemic heart disease, and diabetes mellitus to CCBs. Thus, we named occurrence of at least one of these diseases as "cardiometabolic diseases" and utilise it as a transient confounder for the second set of semi-synthetic data experiments. For confounding intensity, we chose $\beta$ values: [1, 5, 10] and [25, 50, 75] for experiments with sex and cardiometabolic disease as confounder respectively totalling six experiments on semi-synthetic data. In sum, with this confounding generation method, model confounding adjustment ability will be tested with two forms of confounding at multiple $\beta$ intensities.

On the confounding dataset with cardiometabolic disease confounding at level=75, we additionally conducted finite-sample causal estimation experiments. Since estimators for finite-sample estimation are known to be unstable in many cases (e.g. inverse probability weight based estimators) despite asymptotic guarantees[30], we wished to assess our model for finite-sample estimation ability. Thus, we investigated the estimation ability of our proposed model and other deep learning models by applying the models on random sub-samples of this dataset: 2.5%, 5%, 10%, 25%, 50%, and finally, the entire dataset.

### *C. Feature Selection and Pre-Processing*

The modalities of CPRD considered for modelling were sex, region, diagnoses from both primary and secondary care, medications, systolic blood pressure (BP) measurements, and smoking status – marked chronologically across patient timelines.

We mapped Read codes from primary care and ICD-10 codes from secondary care to 1,471 unique ICD-10 diagnostic codes[31], [32] to homogenize disease codes in the dataset; unmapped codes were included for completion. Furthermore, we mapped medication codes to 426 codes in the British National Formulary (BNF)[33] coding format. Systolic BP measurements were grouped into 16 categories based on pre-specified boundaries ([90-116], (116,121], (121,126], … (181,186], >186). Furthermore, we utilised calendar year and age information for the temporal modalities. Each patient $p$ had $n_p$ encounters, or instances of modalities: diagnoses, medications, and systolic BP measurements. Smoking status at baseline, region, and sex were static variables included in modelling.

### D. Proposed Model Development

Our model, T-BEHRT, utilises a modified BEHRT extractor to capture both static and temporal medical history variables and captures initial estimates of RR. Downstream, we use CV-TMLE to correct for bias in initial RR estimate and compute corrected RR (**Fig. 3**).

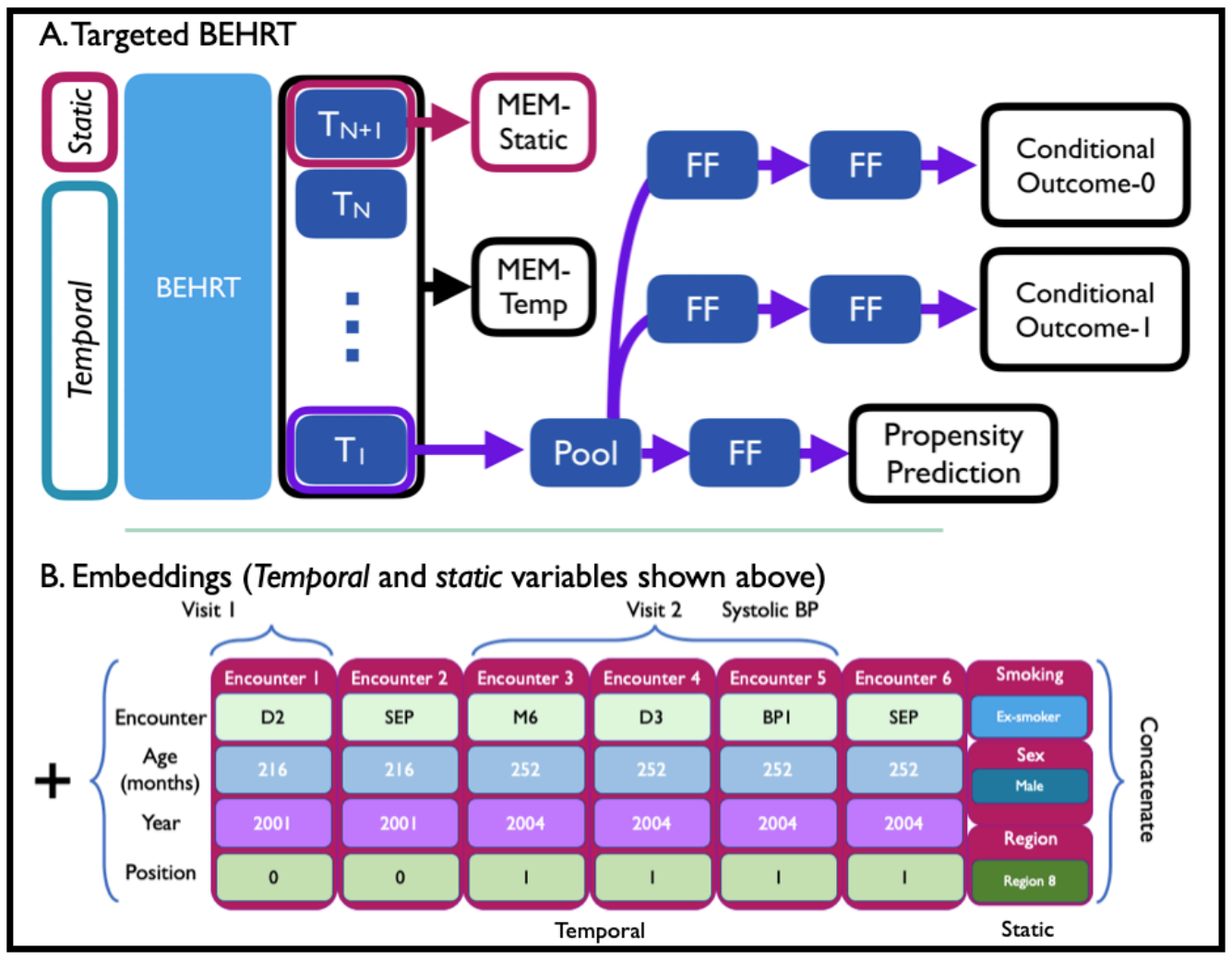

Fig. 3. Targeted BEHRT and Embedding Structure. A. Above, the model is shown. Generally, an input x (static and temporal variables) is fed to a feature extractor, which outputs a dense latent state (for EHR modelling, this feature extractor is BEHRT). The output of the final layer of the BEHRT feature extractor is fed to the Masked EHR modelling (MEM) prediction head to predict any masked encounters. The last token's latent state (TN+1) is fed to a Variational Autoencoder (VAE) neural network to predict masked static variables. The latent state of the first token (T1) is fed to a pooling layer to predict propensity and conditional outcomes with multiple prediction heads. The loss consists of the unsupervised loss from two MEM components – temporal (temp) and static (static) unsupervised data training - and the supervised loss of the propensity and factual outcomes. B. Below, the embedding structure for modelling rich EHR data is shown here. Clinical encounters timestamped by age/year/relative position are converted to vector representations and fed to model as temporal variables. Static data variable embeddings: patient sex, region in UK, and smoking status are concatenated to the temporal variable embeddings.

Intuitively, T-BEHRT first extracts latent EHR features from static covariates and fixed sub-sequences of medical history with BEHRT. Second, the model predicts propensity of exposure and conditional outcome using these grasped features similar to the Dragonnet model. Third, by additionally conducting auxiliary unsupervised learning, the model trains on reconstruction of both static and temporal data with two-part Masked EHR modelling (MEM).

The propensity prediction model is modelled as 1-hidden layer multilayer perceptron (MLP) and for each conditional outcome, we use a 2-hidden layer MLP with Exponential Linear Unit (ELU) activation.

With parameters $\theta$, propensity prediction head $g(x_i)$, and conditional outcome prediction heads, $H(x_i, t_i)$ for input $x_i$ and exposure $t_i$ for patient $i$, the loss:

$$\hat{O}(x_i; \theta) = CrossEntropy(H(x_i, t_i; \theta), y_i) + CrossEntropy(g(x_i; \theta), t_i)$$

Next, we conduct MEM for two-part unsupervised learning: (1) temporal variable and (2) static variable modelling. The first part – unsupervised learning on temporal data – functions similarly to masked language modelling (MLM) in Natural Language Processing.[34] In MLM, the model receives a combination of masked, replaced, and unperturbed tokens (temporal or textual data) and the task is to predict the masked or replaced encounters. We do the same but additionally enforce another constraint: when replacing encounters, we do not replace encounters with those that define the exposure or outcome - antihypertensives and cancer in the current set of experiments. With encounter $j$ for patient $i$ represented as $E_{i,j}$, masked/replaced encounters represented as $\tilde{E}_{i,j}$, BEHRT feature extractor B, temporal unsupervised prediction network M, neural network parameters $\phi_{MEM-Temp}$, we develop objective function:

$$\widehat{\mathcal{L}_{MEM-Temp}}(E_{i,j}; \phi_{MEM-Temp}) = \sum_{j=1}^{|E_i|} CrossEntropy\left(M\left(B(\tilde{E}_{i,j};\ \phi_{MEM-Temp})\right), E_{i,j}\right)$$

For the second part of the MEM, static data modelling, we chose using VAE for unsupervised learning due to cumulative literature empirically demonstrating its strength in representation learning in addition to the utilisation of VAE structures in other causal deep learning models such as CEVAE.[18] As stated, we model region, smoking status at baseline, and sex as static variables; the variables are embedded in high dimensional categorical embeddings and the information is concatenated to the BEHRT encounter embeddings. Thus, the BEHRT model functions as feature extractor for temporal variables and encoder for the VAE. The temporal variables interact with the static variables through the multi-head self-attention mechanism of the BEHRT architecture.[5] For training the VAE, similarly to the temporal modelling, we mask some variables as input, and use a variable-specific decoder to decode the variable (if masked). Specifically, for static variable $x_{v,i}$ of a total of $V$ static variables patient $i$, $q_{\phi_{Enc}}(z_i|x_i)$ representing the encoder, and $p_{\phi_{Dec}}(x_{v,i}|z_i)$ representing the multivariate Bernoulli decoder for variable $v$, the VAE loss is:

$$\widehat{\mathcal{L}_{MEM-Static}}(x_i; \phi_{Enc}, \phi_{Dec}) = \sum_{v=1}^{V}\sum_{i=1}^{n} \log\, p_{\phi_{Dec}}(x_{v,i}|z_i) - \sum_{i=1}^{n} D_{KL}\left(q_{\phi_{Enc}}(z_i|x_i)\, ||p_{\phi_{Dec}}(z_i)\right)$$

The complete objective function to be minimized is:

$$\hat{\theta}, \hat{\varepsilon}, \hat{\phi}_{Enc}, \hat{\phi}_{Dec}, \hat{\phi}_{MEM-Temp} = \underset{\theta,\varepsilon,\phi_{Enc},\phi_{Dec},\phi_{U}}{\text{argmin}} \sum_{i=1}^{n} \hat{O}(x_i;\theta) + \delta\left(\widehat{\mathcal{L}_{MEM-Temp}}(E_{i,j};\phi_{MEM-Temp}) + \widehat{\mathcal{L}_{MEM-Static}}(x_i;\phi_{Enc},\phi_{Dec})\right)$$

With hyperparameter $\delta$ for weighting the contribution of the unsupervised MEM loss terms.

## *E. Benchmarks and causal estimation*

Before pursuing the causal investigations with deep learning modelling, we pre-trained contextualised EHR embeddings and network weights through MEM on the pretraining dataset. This MEM task generally trains weights on all patients in CPRD before progressing to causal modelling (6,777,845 patients in **Fig. 1**).

For the semi-synthetic investigations and first routine clinical investigation, we implemented statistical and deep learning models to serve as benchmarked comparison models for causal inference. The benchmarks include Bayesian Additive Regression Trees (BART)[35], Logistic Regression (LR) and L1/L2 regularization variants, and logistic regression with Targeted Maximum Likelihood Estimation (TMLE).[36] We chose the covariates for these models to be baseline age, smoking status, sex, region, incidence of 33 curated disease groups, and additionally prescription of four additional medications groups. While inclusion of baseline variables in epidemiological observational studies is standard practice, we specifically include the disease and medication groups to enable a fairer comparison to deep learning modelling. Furthermore, diagnoses and medications are known to be confounders in observational studies, so adjustment of these variables is important for causal estimation. To ensure that the diagnoses and medication groups are medically valid clusters of diseases and medications respectively, we utilised groups compiled by past medical research[26], [27]. A deeper explication of statistical model development is given in **Supplementary Methods: Statistical model development**.

To serve as deep learning benchmarks, we implemented staple models in causal deep learning literature for estimating average exposure effect: TARNET, TARNET + MEM (i.e. with unsupervised MEM component), and Dragonnet with BEHRT feature extractor and the embedding format presented in **Fig. 3A**. We initialised these models with pretrained weights. After implementing and evaluating benchmarks, we implemented T-BEHRT with pre-trained network weights where applicable and pursue modelling of semi-synthetic data investigations.

For the semi-synthetic data experiments, we did not feed variables denoting cardiometabolic disease and sex respectively as input; we wish the statistical and deep learning models to infer confounding from remaining input variables. In routine clinical data, the observational studies would often not have access to all confounding variables - thus, important to test models' ability to adjust for confounding given limited input variables.

For all investigations, we conducted experiments with five-fold cross validation causal estimation. We calculated RR on the test dataset for each fold as advised by Chernozhukov et al[16] and compute 95% Confidence Intervals (CI) over the five folds. We computed RR defined by naïve estimator on a finite sample: $\hat{\psi} = \mathbb{E}\left[\frac{\mathbb{E}[\hat{H}(X,1)]}{\mathbb{E}[\hat{H}(X,0)]}\right]$ for TARNET, TARNET-MEM, LR (and L1/L2 regularization variants), and BART. For Dragonnet and T-BEHRT, we use CV-TMLE for estimation of RR. For more information on the CV-TMLE method, advantages over TMLE, and implementation, please refer to Supplementary Methods: CV-TMLE. For models that utilised predicted propensity scores, we conducted propensity score trimming and exclude patients with predicted propensity score greater than 0.97 and less than 0.03[37] before pursuing RR calculation.

We identified the superior model by identifying the model with least Sum Absolute Error (SAE) over the three Beta values for each confounding experiment. We give the Standard Error (SE) for the SAE; this was calculated using additive propagation of error.[38] For deep learning models, we also demonstrate change of SAE as modular additions are incorporated in the model.

## *F. Implementation*

We developed all statistical and deep learning models on python. The deep learning models were implemented with Pytorch[39] Hyperparameters for the BEHRT feature extractor are found in **Supplementary Table 2.** For training all deep learning models, we used the Adam optimizer[40] with exponential decay scheduler (decay rate=0.95) to ensure training convergence. For TARNET-MEM and T-BEHRT, we pre-trained 5 epochs on exclusively the MEM task before initiating joint MEM-causal task training.

# III. Results

## *A. Population Statistics*

TABLE I
POPULATION STATISTICS

| | Classes of antihypertensives | | | | |
|---|---|---|---|---|---|
| | **ACEIs** | **BBs** | **CCBs** | **Diuretics** | **ARBs** |
| **No. (%)** | 186709 (36) | 150098 (29) | 128597 (24) | 28991 (5) | 21970 (4) |
| **Male (%)** | 101629 (54) | 67794 (45) | 60395 (46) | 8134 (28) | 10454 (47) |
| **Smoker (current/ex) (%)** | 101629 (54) | 67794 (45) | 60395 (46) | 8134 (28) | 10454 (47) |
| **YOB (SD)** | 1938 (15) | 1941 (15) | 1936 (14) | 1934 (16) | 1940 (14) |
| **Baseline Age (SD)** | 63 (14) | 59 (14) | 64 (13) | 63 (15) | 63 (13) |
| **Number of visits (SD)** | 7 (4) | 6 (4) | 6 (4) | 4 (4) | 7 (4) |
| **Baseline Year (SD)** | 2001 (4.2) | 1999 (4.3) | 2000 (4.9) | 1996 (5.2) | 2002 (2.8) |

BBs: beta blockers; CCBs: calcium channel blockers; ACEIs: angiotensin-converting-enzyme inhibitors; ARBs: angiotensin receptor blockers; No. :number; YOB: year of birth; baseline: the time of exposure assignment; SD: standard deviation; %: percentage

In the dataset for the investigation of antihypertensives on incident cancer, we identified 186,709, 150,098, 128,597, 28,991, and 21,970 patients for ACEIs, BBs, CCBs, diuretics, ARBs respectively totalling 516,365 patients. We demonstrate population statistics in **Table 1**. Cancer incidence counts/percentage of exposure group were 13,728 /7%, 9,819/7%, 10,232/8%, 1,784/6%, and 1,709/8% for ACEIs, BBs, CCBs, diuretics, ARBs respectively.

## *B. Semi-synthetic Data Experiments*

In the semi-synthetic experiments on confounders cardiometabolic diseases and sex, we tested the T-BEHRT models against several statistical and deep learning benchmarks. In **Fig. 4A/B**, we show SAE with SE measures calculated over all $\beta$-specific semi-synthetic data experiments. We include more detailed experimental results in **Supplementary Table 3**.

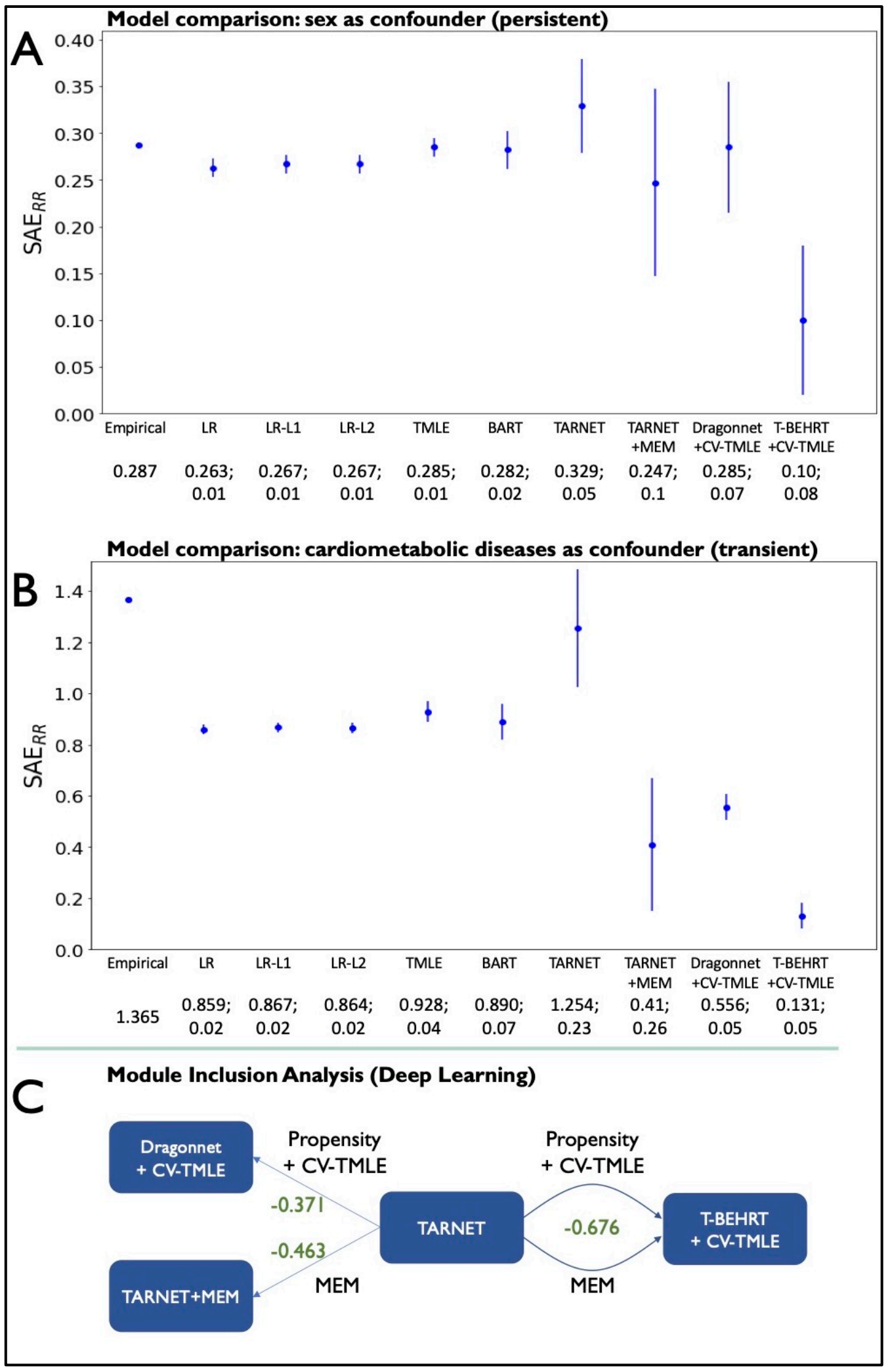

Fig. 4. Experiments on semi-synthetic data with sex (A) and cardiometabolic disease (B) as confounders; module inclusion analysis of causal modules (C). We show Sum Absolute Error (SAE) between ground truth risk ratio (RR) and estimated RR with standard error measures in both panels. The x axis is shown by the models implemented on these datasets, and the y axis is the SAE (lower is better). We present the numerical value and standard error measures underneath the model names. In C, we present the transformation from TARNET into other benchmarks and our proposed model (deep learning models). We show change in average SAE across experiments of transient and persistent confounding in green as model transforms by including varying modules.

We found that our proposed model, T-BEHRT, outperforms all given deep learning and statistical model solutions in terms of SAE whilst maintaining narrow SE as shown in panels A and B (**Fig. 4**). Additionally, across both experiments, we show that deep learning models for EHR benefit from inclusion of propensity score modelling. This is seen by superior performance of both Dragonnet and T-BEHRT in comparison with TARNET, which does not require propensity score modelling. However, by investigating the inclusion of various modules appended to the chassis of TARNET shown in our module inclusion analysis (**Fig. 4C)**, we see that inclusion of MEM might improve RR estimation in a parallel way; the TARNET model with inclusion of MEM (SAE reduction of 0.463) does approximately as well as Dragonnet + CV-TMLE (SAE reduction of 0.371) averaged over experiments of persistent and transient confounders. Regardless, the improvement in combining both MEM and propensity modelling and forming T-BEHRT demonstrates greatest SAE reduction of 0.676.

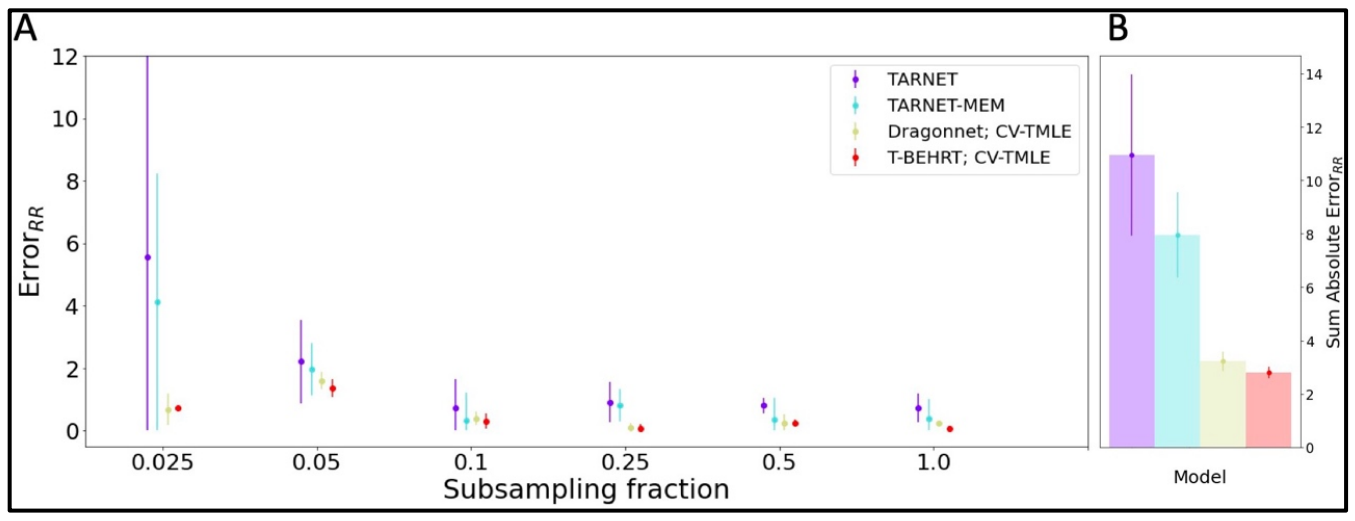

Fig. 5. Finite-sample experiments on semi-synthetic data. A. We conduct experiments on finite subsamples of the semi-synthetic dataset for cardiometabolic confounding (Beta 75). The subsampling fraction of the dataset is shown on the x axis. The y axis shows error from ground truth risk ratio (RR). The models: TARNET (and with Masked EHR Modelling), Dragonnet + CV-TMLE, and T-BEHRT estimate RR on the fractional samples of the dataset. The point estimate is the mean value on five-fold cross validation and the error bars represent 95% confidence intervals for those point estimates of RR. B. Sum Absolute Error (SAE) across the seven subsamples of the dataset are shown for each model (Denoted by colour) is shown. The four models are represented by the four bars with interval defined by Standard Error (SE) and colour scheme is the same as part A.

In the finite-sample estimation experiments shown in **Fig. 5**, we demonstrate T-BEHRT outperforms other models in RR estimation in individual and across data subsamples. While improvement of T-BEHRT over Dragonnet is less pronounced than over other models, panel B shows that T-BEHRT still demonstrates superior RR estimation performance with respect to the deep learning benchmarks. Furthermore, we show that inclusion of MEM aids more precise estimation of RR; TARNET-MEM and T-BEHRT perform better than TARNET and Dragonnet respectively across all sizes of dataset. However, we note the application of CV-TMLE is more important than MEM in smaller datasets as seen by superior performance of Dragonnet + CV-TMLE as opposed to TARNET + MEM. Furthermore, models equipped with CV-TMLE maintain relatively stable SAE across subsampling fractions while TARNET and derivatives suffer in RR estimation in smaller datasets. Lastly, we see as dataset size increases, SAE across models begin to converge. Theoretically, as the number of samples increases, we would be slowly mitigating the finite-sample bias, and thus, the performance of TARNET and derivatives should be similar to those of models assisted by propensity modelling also noted by Shi et al[7].

We apply our model on the routine clinical data study of effect of ACEIs on incident cancer with respect to other antihypertensive drug classes and demonstrate the results in **Fig. 5**. Across all four drug class comparisons, while the empirical RR often tends away from null implying a preventive or harmful effect, we show that our model's 95% confidence interval for RR covers the null hypothesis (1.0 RR) across almost all drug class comparisons with exception of CCBs.

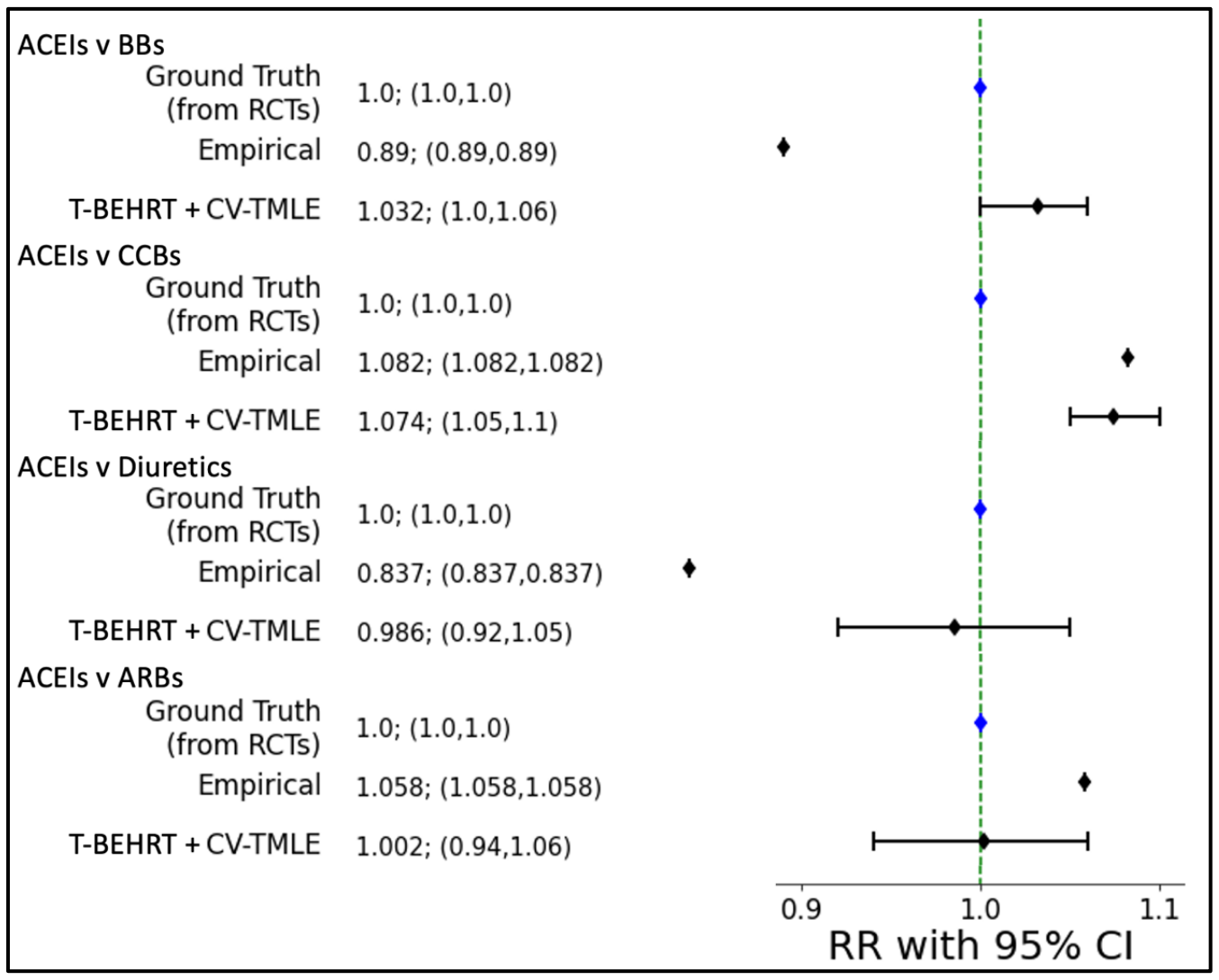

Fig. 6. Application of T-BEHRT on routine clinical data: Effect of ACEI on incident cancer with respect to BBs, CCBs, Diuretics, and ARBs. This forest plot has four parts; one for each of the antihypertensive groups. We show CV-TMLE risk ratio (RR) estimates with 95% Confidence Intervals (CI) on our T-BEHRT model. In addition, we show empirical RR in the observational cohort selected for these experiments. The ground truth is assumed to be 1.0 (null) validated by multiple RCTs. BBs: beta blockers; CCBs: calcium channel blockers; ACEIs: angiotensin-converting-enzyme inhibitors; ARBs: angiotensin receptor blockers; RR: risk ratio

## IV. Discussion

In this paper, by utilising large scale comprehensive EHR and deep learning methods, we have developed a causal model for observational causal inference for clinical sciences. We have validated our model against benchmarks across six semi-synthetic experiments in addition to a finite-sample estimation study and found T-BEHRT to demonstrate superior estimation in the cases of persistent and transient confounding. Finally, we apply our model to a routine clinical data observational study.

Our work has contributions to the field of EHR based deep learning research. First, the model consolidates multiple static and temporal data embeddings into a unified embedding structure thus allowing adjustment over multiple datatypes in rich EHR. Second, T-BEHRT conducts novel MEM unsupervised learning using MLM and VAE-based representation learning in tandem with the causal inference objective. We demonstrate the benefits of unsupervised learning in the context of average RR estimation in multiple experiments as well. In our assessment, this is the first work conducting causal inference incorporating unsupervised learning on multiple data types. Third, we introduce CV-TMLE estimation correction for less biased RR estimation for deep learning causal models on EHR data. While the utility of propensity modelling and CV-TMLE is as effective as MEM modelling for RR estimation in larger dataset sizes, we found that in our finite-sample estimation experiments, the former is far more critical for accurate RR estimation. Finally, in our observational study, we show that our model can be applied to test a clinical hypothesis in an observational setting.

Our work has some limitations and scope to grow as well. First and most fundamentally, we note that EHR data may not completely capture confounding variables in our observational studies. Latent confounding can be investigated in future works with latent variable modelling techniques[8] to rich EHR. Furthermore, we note that we have included the data modalities of diagnoses, medications, smoking, sex, and systolic BP; however, better confounding adjustment might manifest with full utilisation of the modalities that rich databases like CPRD have to offer. In this work, we have included sex as an explicit feature in deep learning modelling to adhere to conventions of adjustment in clinical/epidemiological research; however we note, it is shown that BEHRT naturally captures sex[5] without explicit inclusion as a variable. Also, in terms of data curation, we have allocated patients into an exposure group based on first prescription of class of antihypertensives. Subgroup investigations stratified by intensity and duration of drug class should be additionally pursued in future studies. Lastly, T-BEHRT estimated null in most drug comparisons in the routine clinical data study, but we note that our model finds the comparison to CCBs to deviate from the null. While findings from the RCTs generally demonstrate that antihypertensives have null effect on cancer, the evidence regarding CCBs is still conflicting and further research is required[11]. Lastly, in our modelling, we cannot entirely rule out residual confounding. A variety of variables may be affecting outcome may be unadjusted for (explicitly or through latent representation modelling) and as discussed, further modality inclusion is necessary in future work.

## V. Conclusion

To conclude, we have developed a deep learning model for EHR data for superior and reliable estimation of RR. T-BEHRT has performed optimally in semi-synthetic data experiments with both, persistent and transient confounding and can be applied to an observational study on routine clinical data. Thus, in the future, this model should be further tested and applied to investigate other causal hypotheses questions using routine EHR.

## Appendix

### *A. Supplementary Methods: Semi-synthetic data simulation*

Data generation of sequential, temporal variables is a difficult task, and currently, there is no medically validated method of generating realistic EHR medical history, exposure assignment, and outcome. Thus, we limit data generation to factual/counterfactual outcome generation and utilise routine clinical data components: (1) medical history with balanced exposure groups and (2) exposure status for an observational investigation of an association.

In order to create this semi-synthetic dataset, we first form the dataset for the investigation: effect of antihypertensives on incident cancer allowing us access to components (1) and (2). Since outcome generation using only (2), exposure status would be unconfounded and thus too rudimentary a function to test the tolerance of models, we force the relationship between treatment and outcome to be confounded. Since confounding often manifests partly due to imbalanced variables between exposure groups, we find an imbalanced variable, $Z_i$ in (1), routine medical history. We then force this imbalanced variable, $Z_i$ to be a confounder and generate conditional outcome from a sampling function:

$$y_i = Bernoulli\left(\sigma\left(at_i + m\beta(\lambda_i + c)\right)\right)$$

$\lambda_i$ represents $P(T_i|Z_i)$, $T_i$ is the exposure for patient *i*, $y_i$ is the outcome for patient *i*, $\sigma$ is the sigmoid function, and $\beta$, the magnitude of confounding. Variables *a, m,* and *c* are coefficients to terms weighting their importance in the function. Intuitively, we first model the association between a variable $Z_i$ and exposure ($P(T_i|Z_i)$) with $\lambda_i$. Next, we generate $y_i$ with two variables: the variable $Z_i$ and exposure ($P(Y_i|T_i, Z_i)$). In this way, we form an association between $Z_i$ and exposure and $Z_i$ and the outcome; with association to both, we force $Z_i$ to be a confounder in this data generating process. This process synthesizes controlled confounded observational data; by generating the outcome with this function, we control confounding with a confounder $Z_i$. Thus, in this way we can generate factual/counterfactual outcomes and consequently ground truth RR. Lastly, we can modify $\beta$ value to vary the degree of confounding in the data generation process.

### B. Supplementary Methods: CV-TMLE

After using T-BEHRT to compute initial estimates, we use CV-TMLE[41] for correction of these estimates. We refer readers to the source material for theory behind TMLE and the cross validated form: CV-TMLE[17], [41]. In brief, the original formulation of the CV-TMLE algorithm requires *k* targeting steps for each of the *k* folds for each of the iterations pre-defined in the iterative version of TMLE. However, Levy forms a simpler construction of the CV-TMLE which is less computationally cumbersome; the advised method is to pool all the initial estimates across folds and compute corrected estimates vis-à-vis a standard TMLE update step[41]. Albeit procedurally different, the original formulation and Levy's more recent formulation of CV-TMLE are identical mathematically and in function. According to our research, this is the first work utilising CV-TMLE paired with deep learning methods.

CV-TMLE provides a host of benefits to observational causal inference. As recommended, by Chernozhukov[42], with validation in k-fold framework, CV-TMLE is a form of TMLE which is robust to issues of fold-wise overfitting whilst conducting k-fold cross validation[41]. Furthermore, previous works show that the CV-TMLE estimator provides more robustness than other cross-validated estimators (e.g. CV-AIPTW) in the case of violations of the assumption of overlap[43].

### C. Supplementary Methods: Statistical Model Development

In statistics models, we used RR for estimation of causal effect. The covariates adjusted for were: baseline age (continuous variable: [0,1]), sex (male/female), region, smoking status (smoker/non-smoker), chronic kidney disease (yes/no), human immunodeficiency virus/acquired immune deficiency syndrome (yes/no), ischaemic heart disease (yes/no), cardiac arrhythmia (yes/no), stroke (yes/no), heart failure (yes/no), anaemia (yes/no), diabetes mellitus (yes/no), hypertension (yes/no), osteoporosis (yes/no), arthritis (yes/no), connective tissue disorder (yes/no), gout (yes/no), rheumatoid arthritis (yes/no), peptic ulcer disease (yes/no), liver disease (yes/no), asthma (yes/no), peripheral arterial disease (yes/no), chronic obstructive pulmonary disorder (yes/no), hemiplegia (yes/no), epilepsy (yes/no), dementia (yes/no), learning disorder (yes/no), eating disorder (yes/no), adjustment (yes/no), anxiety (yes/no), affective disorder (yes/no), depression (yes/no), bipolar disorder (yes/no), psychoses (yes/no), schizophrenia (yes/no), hyperlipidaemia (yes/no), obesity (yes/no), substance abuse (yes/no), anticholinergics (yes/no), drugs that cause gastrointestinal bleedings (yes/no), statins (yes/no), drugs for diabetes (yes/no). The exposure variable was antihypertensive medications (class 1/class 2). The outcome was defined as the synthetic outcome/cancer (yes/no). The models were fit and tested using five-fold validation. The naïve risk ratio estimates were calculated on the testing dataset in each fold and mean risk ratio (RR) estimate and 95% confidence intervals (CI) for estimates were derived.

The TMLE was developed using two logistic regression models – one for outcome prediction and the other for exposure prediction. The outcome prediction model adjusted for covariates and exposure variable listed above, and the exposure prediction model used just the covariates. The TMLE algorithm was fit and tested using five-fold validation. The TMLE RR estimates were calculated on the testing dataset in each fold and mean RR estimate and 95% (CI) for estimates were derived.

### D. Supplementary Methods: Codes for Cancer

Provided in Supplementary Table 1, we have ICD-10$^{th}$ revision codes for cancer stratified by type.

| Supplementary Table I<br>ICD-10$^{TH}$ Revision codes for cancer |
|---|
| Other cancer: C79, D07, D38, D39, C71, D37, C73, C72, C75, C74, C76, C00, C39, C38, C34, D05, C26, C37, C14, C97, C69, R18, C70, C44, C45, C46, C47, C40, C41, C43, D32, C48, C49, D09, D33, D49, D48, D18, D43, D42, D41, D40, D47, D44, C80, B21, E26, C78 |
| Leukaemia: C93, C92, C91, C90, C95, C86, C94, C88, D47, D46, D45 |
| Urinary cancers: C63.9, C67.6, C68.5, C66.4, C66.2, C66.8, C66.0, C68.6, D09.1, C66.5, C68.7, C66.1, C68.1, C66.3, C68.4, C68.3, C66.9, C66.7, C66.6, C68.2, C68.8, C66.X, C68.0, C80.1, C68.9, C67.7 |
| Bladder cancer: C67.9, C67.2, C67.0, C67.1, C67.5, C67.3, C67.8, C67.6, C67.4, D09.0, C67.7 |
| Renal cancer: C65.2, C64.7, C64.X, C65.7, C65.9, C65.4, C65.0, C65.8, C65.X, C64.6, C65.1, C64.3, C64.0, C64.4, C65.5, C65.3, C64.8, C64.5, C64.2, C64.1, C64.9, C65.6 |
| Male reproductive cancers: C63.3, C60.4, C63.2, C60.1, C62.3, C63.9, C63.7, C63.1, C60.2, C62.1, C62.8, C62.0, C63.5, C62.2, C62.6, C62.7, D07.6, C63.8, C63.4, C60.3, C60.6, C60.8, C60.0, C60.5, D07.4, C60.7, C60.9, C62.4, C63.0, C62.9, C63.6, C62.5 |
| Prostate cancer: C61.9, C61.6, C61.2, C61.5, C61.3, C61.7, C61.4, D07.5, C61.8, C61.1, C61.0, C61.X |
| Female reproductive cancers: C57, D07, C55, C54, C53, C52, C51, D25, C58, D49 |
| Ovarian cancer: C56.1, C57.8, C56.8, C56.9, C56.X, C56.5, C56.3, C56.4, C56.6, C56.7, D63.0, C56.2, C56.0 |
| Cervical cancer: D06.4, D06.2, D06.9, D06.1, C53.8, C53.9, C53.2, D06.7, C53.5, D06.8, C53.1, D06.0, D06.3, C53.6, C53.4, D06.6, C53.3, D06.5, C53.7, C53.0 |
| Breast cancer: D05.6, C50.9, C50.5, D05.7, D05.4, D05.3, D05.2, D05.8, C50.1, C50.4, D05.5, C50.2, D05.1, D05.0, C50.8, C50.3, D05.9, C50.7, C50.0, C43.5, C50.6 |
| Skin cancer: D03.7, D04.4, C44.6, D22.1, D04.3, D22.7, C79.2, C44.0, D04.0, C44.9, C44.5, D04.7, D03.0, D22.5, D03.8, C44.4, D04.8, D22.4, D03.5, C44.3, D22.3, D22.9, D22.6, D49.2, C43.8, C60.0, C44.8, D22.2, D03.2, C44.1, D04.2, D03.4, D04.9, D04.1, D03.6, C46.0, D22.8, D22.0, C44.2, C43.9, C44.7, D03.1, D04.6, D03.3, D03.9, D04.5 |
| Gastrointestinal cancers: C22, C23, C21, C26, D01, C24, D49, C18, C17 |
| Respiratory cancers: C33.7, D02.6, C34.8, C34.0, C76.1, D02.0, D02.2, D02.1, C38.4, D02.8, C45.0, C39.0, C33.8, C33.2, C34.9, D02.9, D02.3, C45.9, C33.9, C33.4, C34.3, D49.1, C33.1, C33.5, C33.6, C33.0, C34.1, C39.8, C34.2, C39.9, C33.X, D02.5, C45.7, C33.3, D02.4, D02.7 |
| Unspecified cancer: D01.2, Z92.8, C46.9, D01.8, D01.9, D09.9, C97.X, Z08.8, D01.0, D49.2, C80.9, D49.9, B20.0, D01.5, Z08.9, D01.1, D01.4, D01.7, D01.6, D01.3, K63.5, D49.5, C80.1, D49.7 |
| Lung cancer: C34.3, C34.4, C34.0, C34.8, C34.9, C34.7, C34.5, C34.1, D02.2, C34.2, C34.6 |
| Head and neck cancers: C79, C39, C13, C12, C11, C10, C31, C30, C32, D00, C14, C46, D02, C43, C49, C08, C09, C00, C01, C02, C03, C04, C05, C06, C07 |
| Pancreatic cancer: C25.3, D49.0, C25.5, C25.4, C25.1, C25.6, C25.9, C25.7, C25.8, C25.0, C25.2 |
| Liver cancer: C22.3, C22.9, C22.6, D01.5, C22.7, C22.4, C22.2, C22.1, C22.5, C22.8, C22.0 |
| Stomach cancer: C16.3, D00.2, C16.4, C16.8, C16.5, C16.7, C16.2, C16.1, C16.0, C16.9, C16.6 |
| Oesophageal cancer: C15.4, C15.7, C15.2, C15.6, C15.8, C15.9, D00.1, C15.0, C15.1, C15.5, C15.3 |
| Rectal cancer: C20.4, C20.8, D01.2, C19.8, C19.9, C20.2, C19.0, C21.8, C20.1, C20.5, C20.3, C19.6, C19.X, C19.5, C20.9, C20.X, D01.1, C19.4, C19.7, C20.6, C19.2, C20.0, C20.7, C19.3, C19.1 |
| Colon cancer: C18.5, C18.7, C18.4, C18.3, C18.6, C18.1, C18.2, C18.0, D01.0, C18.8, C18.9 |
| Lymphatic cancer: C81, C82, C90, C84, C85, C88, C91, B21, C83, D47, C96 |
| Metastatic cancer: C79.3, C79.7, C77.1, G89.3, C79.2, M84.5, C77.8, C78.4, C77.9, C78.0, C78.7, C78.6, C79.1, C78.9, C79.0, C79.5, C78.1, C78.8, C77.5, C77.4, C79.9, C77.3, C79.6, C79.4, C77.7, C77.2, C79.8, C78.5, C43.9, C78.2, C77.6, C78.3, C77.0 |

The codes considered for incident cancer is shown in the table. The codes are stratified by type of cancer.

## E. *Supplementary Methods: Hyperparameters for deep learning models*

Provided in Supplementary Table 2, we have hyperparameters used for the BEHRT architecture models.

SUPPLEMENTARY TABLE 2
HYPERPARAMETERS FOR BEHRT CHASSIS

| **Hyperparameter** | **Attribute** |
|---|---|
| Hidden BEHRT | 150 |
| Intermediate BEHRT Layer size | 108 |
| Hidden dropout probability | 0.3 |
| Attention dropout probability | 0.4 |
| Number of hidden layers (BEHRT) | 4 |
| Hidden activation functions | GeLU |
| Initialiser range of parameters | 0.02 |
| $n$ | 200 |
| $\delta$ | 0.1 |

## F. *Supplementary Methods: Hyperparameters for deep learning models*

Provided in Supplementary Table 3, we have the raw risk ratio estimates across the semi-synthetic data experiments.

SUPPLEMENTARY TABLE 3
RISK RATIO ESTIMATES ACROSS SEMI-SYNTHETIC DATA EXPERIMENTS

| | | | **Risk ratio** | | **Error** |
|---|---|---|---|---|---|
| **Cardio-metabolic disease** | **Beta** | **25** | **50** | **75** | |
| | **Modelling** | | | | |
| | Ground Truth | 2.207 | 2.727 | 3.178 | 1.555 |
| | Empirical | 2.532 | 3.251 | 3.883 | 1.365 |
| Statistical | LR | 2.398; (2.37, 2.43) | 3.003; (2.97, 3.03) | 3.569; (3.5, 3.64) | 0.859; 0.02 |
| | LR-L1 | 2.399; (2.37, 2.43) | 3.005; (2.97, 3.04) | 3.576; (3.51, 3.64) | 0.867; 0.02 |
| | LR-L2 | 2.399; (2.37, 2.43) | 3.004; (2.97, 3.03) | 3.574; (3.5, 3.64) | 0.864; 0.02 |
| | TMLE | 2.411; (2.28, 2.54) | 3.005; (2.87, 3.14) | 3.622; (3.4, 3.84) | 0.928; 0.04 |
| | BART | 2.398; (2.37, 2.43) | 3.011; (2.98, 3.04) | 3.592; (3.53, 3.65) | 0.890; 0.07 |
| Deep learning | TARNET | 2.283; (2.21, 2.35) | 3.183; (2.76, 3.6) | 3.899; (3.43, 4.36) | 1.254; 0.23 |
| | TARNET + MEM | 2.226; (2.14, 2.31) | 2.719; (2.35, 3.09) | 3.561; (2.94, 4.19) | 0.41; 0.26 |
| | Dragonnet +CV-TMLE | 2.281; (2.26, 2.31) | 2.954; (2.91, 3.0) | 2.922; (2.85, 2.99) | 0.556; 0.05 |
| | **T-BEHRT+ CV-TMLE** | 2.263; (2.24, 2.29) | 2.753; (2.71, 2.8) | 3.227; (3.14, 3.31) | 0.131; 0.05 |
| | | | | | |
| | | | | | |
| | | | | | |
| | | **Risk ratio** | **Error** | | |
| **Sex** | **Beta** | **1** | **5** | **10** | |
| | **Modelling** | | | | |
| | Ground Truth | 1.465 | 1.926 | 2.154 | |
| | Empirical | 1.456 | 1.823 | 1.979 | 0.287 |
| Statistical | LR | 1.455; (1.45, 1.46) | 1.83; (1.81, 1.85) | 1.996; (1.95, 2.04) | 0.263; 0.01 |
| | LR-L1 | 1.455; (1.45, 1.46) | 1.83; (1.81, 1.85) | 1.992; (1.95, 2.04) | 0.267; 0.01 |
| | LR-L2 | 1.455; (1.45, 1.46) | 1.829; (1.81, 1.85) | 1.993; (1.95, 2.04) | 0.267; 0.01 |
| | TMLE | 1.453; (1.44, 1.47) | 1.824; (1.79, 1.86) | 1.982; (1.93, 2.03) | 0.285; 0.01 |
| | BART | 1.455; (1.45, 1.46) | 1.826; (1.8, 1.85) | 1.981; (1.94, 2.02) | 0.282; 0.02 |
| Deep learning | TARNET+ MEM | 1.457; (1.44, 1.47) | 1.863; (1.77, 1.96) | 1.977; (1.72, 2.24) | 0.247; 0.1 |
| | TARNET | 1.465; (1.44, 1.49) | 1.803; (1.71, 1.89) | 1.948; (1.83, 2.02) | 0.329; 0.05 |
| | Dragonnet +CV-TMLE | 1.469; (1.45, 1.49) | 1.827; (1.78, 1.87) | 1.973; (1.85, 2.09) | 0.285; 0.07 |
| | **T-BEHRT+ CV-TMLE** | **1.47; (1.45, 1.49)** | **1.854; (1.81, 1.9)** | **2.132; (1.98, 2.29)** | **0.1; 0.08** |

This table shows the risk ratio and standard deviation (five-fold) for statistical and deep learning models over the two semi synthetic experiments with cardiometabolic diseases and sex as confounders (top and bottom respectively). Over various values of Beta, confounding experiments are conducted. Ground truth risk ratio is calculated and displayed for both experiments. Risk ratio and 95% confidence interval for each model is presented in the table. The sum absolute error from ground truth risk ratios for models over all the confounding experiments and 95% confidence interval is shown in the far-right column. Bolded models are best statistical and deep learning models. LR: Logistic Regression; LR-L1; Logistic Regression with L1 penalty; LR-L2; Logistic Regression with L2 penalty; TMLE: Targeted Maximum Likelihood Estimation; BART: Bayesian Additive Regression Trees; T-BEHRT: Targeted BEHRT

## Acknowledgment

We thank Professors Mark van der Laan and Victor Veitch for constructive comments and discussion during the formative portions of this work.

## References and Footnotes

### A. References